\newif\iftwocol\twocoltrue              

\documentclass[twoside,twocolumn]{article}
\usepackage{latexsym,amsmath,amssymb,hyperref}
\def\eqvsp{}  \newdimen\paravsp  \paravsp=1.3ex
\makeatletter
\def\section{\@startsection {section}{1}{\z@}{-2.0ex plus
    -0.5ex minus -.2ex}{1.5ex plus 0.3ex minus .2ex}{\large\bf\raggedright}}
\makeatother
\def\,{\mskip 3mu} \def\>{\mskip 4mu plus 2mu minus 4mu} \def\;{\mskip 5mu plus 5mu} \def\!{\mskip-3mu}
\def\dispmuskip{\thinmuskip= 3mu plus 0mu minus 2mu \medmuskip=  4mu plus 2mu minus 2mu \thickmuskip=5mu plus 5mu minus 2mu}
\def\textmuskip{\thinmuskip= 0mu                    \medmuskip=  1mu plus 1mu minus 1mu \thickmuskip=2mu plus 3mu minus 1mu}
\textmuskip
\def\beq{\eqvsp\dispmuskip\begin{equation}}    \def\eeq{\eqvsp\end{equation}\textmuskip}
\def\beqn{\eqvsp\dispmuskip\begin{displaymath}}\def\eeqn{\eqvsp\end{displaymath}\textmuskip}
\def\bqa{\eqvsp\dispmuskip\begin{eqnarray}}    \def\eqa{\eqvsp\end{eqnarray}\textmuskip}
\def\bqan{\eqvsp\dispmuskip\begin{eqnarray*}}  \def\eqan{\eqvsp\end{eqnarray*}\textmuskip}

\def\paradot#1{\vspace{\paravsp plus 0.5\paravsp minus 0.5\paravsp}\noindent{\bf\boldmath{#1.}}}
\def\paranodot#1{\vspace{\paravsp plus 0.5\paravsp minus 0.5\paravsp}\noindent{\bf\boldmath{#1}}}
\def\req#1{(\ref{#1})}

\def\epstr{\epsilon}                    
\def\nq{\hspace{-1em}}
\def\fr#1#2{{\textstyle{#1\over#2}}}

\def\SetR{I\!\!R}
\def\SetN{I\!\!N}

\def\qmbox#1{{\quad\mbox{#1}\quad}}

\def\P{{\rm P}}                         

\def\v{\boldsymbol}

\def\g{\gamma}

\def\t{\theta}

\def\A{{\cal A}}
\def\O{{\cal O}}
\def\R{{\cal R}}
\def\S{{\cal S}}
\def\H{{\cal H}}
\def\X{{\cal X}}
\def\T{{\cal T}}
\def\Agent{\text{Agent}}
\def\Env{\text{Env}}
\def\p{{\scriptscriptstyle+}}
\def\vi{{\scriptscriptstyle\bullet}}
\def\CL{\text{CL}}
\def\CLN{\text{CL}}
\def\Cost{\text{Cost}}

\begin{document}

\title{\vspace{-4ex}
\vskip 2mm\bf\Large\hrule height5pt \vskip 4mm
Feature Markov Decision Processes
\vskip 4mm \hrule height2pt}
\author{{\bf Marcus Hutter}\\[3mm]
\normalsize RSISE$\,$@$\,$ANU and SML$\,$@$\,$NICTA \\
\normalsize Canberra, ACT, 0200, Australia \\
\normalsize \texttt{marcus@hutter1.net \ \  www.hutter1.net}
}
\date{24 December 2008}
\maketitle

\begin{abstract}\bf
General purpose intelligent learning agents cycle through
(complex,non-MDP) sequences of observations, actions, and rewards.
On the other hand, reinforcement learning is well-developed for
small finite state Markov Decision Processes (MDPs).
So far it is an art performed by human designers to extract the
right state representation out of the bare observations, i.e.\ to
reduce the agent setup to the MDP framework.
Before we can think of mechanizing this search for suitable MDPs,
we need a formal objective criterion. The main contribution of this
article is to develop such a criterion.
I also integrate the various parts into one learning algorithm.
Extensions to more realistic dynamic Bayesian networks
are developed in the companion
article \cite{Hutter:09phidbn}.

\end{abstract}

\section{Introduction}\label{secIntro}

\paradot{Background \& motivation}
Artificial General Intelligence (AGI) is concerned with designing
agents that perform well in a wide range of environments
\cite{Goertzel:07,Hutter:07iorx}.
%
Among the well-established ``narrow'' AI approaches, arguably
Reinforcement Learning (RL) pursues most directly the same
goal. RL considers the general agent-environment setup in which an
agent interacts with an environment (acts and observes in cycles)
and receives (occasional) rewards. The agent's objective is to
collect as much reward as possible. Most if not all AI problems
can be formulated in this framework.

The simplest interesting environmental class consists of finite
state fully observable Markov Decision Processes (MDPs)
\cite{Puterman:94,Sutton:98}, which is reasonably well understood.
%
Extensions to continuous states with (non)linear function
approximation \cite{Sutton:98,Gordon:99}, partial observability
(POMDP) \cite{Kaelbling:98,Ross:08pomdp}, structured MDPs (DBNs)
\cite{Strehl:07}, and others have been considered, but the
algorithms are much more brittle.

In any case, a lot of work is still left to the designer, namely to
extract the right state representation (``features'') out of the
bare observations. Even if {\em potentially} useful representations
have been found, it is usually not clear which one will turn out to
be better, except in situations where we already know a perfect
model.
Think of a mobile robot equipped with a camera plunged into an
unknown environment. While we can imagine which image features are
potentially useful, we cannot know which ones will actually be
useful.

\paradot{Main contribution}
Before we can think of mechanically searching for the ``best'' MDP
representation, we need a formal objective criterion. Obviously, at
any point in time, if we want the criterion to be effective it can
only depend on the agents past experience.
The main contribution of this article is to develop such a criterion.
%
Reality is a non-ergodic partially observable uncertain
unknown environment in which acquiring experience can be expensive.
So we want/need to exploit the data (past experience) at hand
optimally, cannot generate virtual samples since the model is not
given (need to be learned itself), and there is no reset-option.
%
In regression and classification, penalized maximum likelihood
criteria \cite[Chp.7]{Hastie:01}
have successfully been used for semi-parametric model selection. It
is far from obvious how to apply them in RL.
%
Ultimately we do not care about the observations but the rewards.
The rewards depend on the states, but the states are arbitrary in
the sense that they are model-dependent functions of the data.
Indeed, our derived Cost function cannot be interpreted as a usual
model+data code length.

\paradot{Relation to other work}
As partly detailed later, the suggested $\Phi$MDP model could be regarded %
as a scaled-down practical instantiation of AIXI \cite{Hutter:04uaibook,Hutter:07aixigentle}, %
as a way to side-step the open problem of learning POMDPs, %
as extending the idea of state-aggregation from planning (based on bi-simulation metrics \cite{Givan:03}) to RL (based on code length), %
as generalizing U-Tree \cite{McCallum:96} to arbitrary features, %
or as an alternative to PSRs \cite{Singh:03} for which proper learning algorithms have yet to be developed. %

\paradot{Notation}
Throughout this article, $\log$ denotes the binary logarithm, %
$\epstr$ the empty string, %
and $\delta_{x,y}=\delta_{xy}=1$ if $x=y$ and $0$ else is the Kronecker symbol. %
I generally omit separating commas if no confusion arises, in
particular in indices.
For any $x$ of suitable type (string,vector,set), I define
string $\v x = x_{1:l} = x_1...x_l$, %
sum $x_\p=\sum_j x_j$, union $x_*=\bigcup_j x_j$, and vector $\v x_\vi=(x_1,...,x_l)$, %
where $j$ ranges over the full range $\{1,...,l\}$
and $l=|x|$ is the length or dimension or size of $x$. %
$\hat x$ denotes an estimate of $x$.
$\P(\cdot)$ denotes a probability over states and rewards or parts
thereof. I do not distinguish between random variables $X$ and
realizations $x$, and abbreviation $\P(x):=\P[X=x]$ never
leads to confusion.
%
More specifically, $m\in\SetN$ denotes the number of states, %
$i\in\{1,...,m\}$ any state index, %
$n\in\SetN$ the current time, %
and $t\in\{1,...,n\}$ any time. %
Further, in order not to get distracted at several places I gloss
over initial conditions or special cases where inessential. Also
0$*$undefined=0$*$infinity:=0.

\section{Feature Markov Decision Process ($\mathbf\Phi$MDP)}\label{secPhiMDP}

This section describes our formal setup. It consists of the
agent-environment framework and maps $\Phi$ from
observation-action-reward histories to MDP states. I call this
arrangement ``Feature MDP'' or short $\Phi$MDP.

\paradot{Agent-environment setup}
I consider the standard agent-environment setup \cite{Russell:03}
in which an {\em Agent} interacts with an {\em Environment}.
The agent can choose from actions $a\in\A$ (e.g.\ limb movements)
and the environment provides (regular) observations $o\in\O$ (e.g.\
camera images) and real-valued rewards $r\in\R\subseteq\SetR$ to the
agent. The reward may be very scarce, e.g.\ just +1 (-1) for winning
(losing) a chess game, and 0 at all other times
\cite[Sec.6.3]{Hutter:04uaibook}.
This happens in cycles $t=1,2,3,...$: At time $t$, after observing
$o_t$, the agent takes action $a_t$ based on
history $h_t:=o_1 a_1 r_1...o_{t-1} a_{t-1} r_{t-1}o_t$.
Thereafter, the agent receives reward $r_t$. Then the next cycle
$t+1$ starts. The agent's objective is to maximize his long-term reward.
Without much loss of generality, I assume that $\A$, $\O$, and $\R$
are finite. Implicitly I assume $\A$ to be small, while $\O$ may
be huge.

The agent and environment may be viewed as a pair or triple of interlocking
functions of the history $\H:=(\O\times\A\times\R)^*\times\O$:
\bqan
  & & \nq\Env:\H\times\A\times\R\leadsto\O, \quad   o_n = \Env(h_{n-1}a_{n-1}r_{n-1}), \\
  & & \nq\!\!\!\!\!\Agent:\H\leadsto\A, \qquad\qquad\quad a_n = \Agent(h_n), \\
  & & \nq\Env:\H\times\A\leadsto\R, \qquad\quad       r_n = \Env(h_n a_n).
\eqan
where $\leadsto$ indicates that mappings $\to$ might be stochastic.

The goal of AI is to design agents that achieve high (expected)
reward over the agent's lifetime.

\paradot{(Un)known environments}
For known \Env(), finding the reward maximizing agent is a
well-defined and formally solvable problem
\cite[Chp.4]{Hutter:04uaibook}, with computational efficiency being
the ``only'' matter of concern. For most real-world AI problems
\Env() is at best partially known.

Narrow AI considers the case where function \Env() is either known
(like in blocks world), or essentially known (like in chess, where
one can safely model the opponent as a perfect minimax player), or
\Env() belongs to a relatively small class of environments (e.g.\
traffic control).

The goal of AGI is to design agents that perform well in a large
range of environments \cite{Hutter:07iorx}, i.e.\ achieve high
reward over their lifetime with as little as possible assumptions
about Env(). A minimal necessary assumption is that the environment
possesses {\em some} structure or pattern.

From real-life experience (and from the examples below) we know that
usually we do not need to know the complete history of events in
order to determine (sufficiently well) what will happen next and to
be able to perform well. Let $\Phi(h)$ be such a ``useful'' summary of
history $h$.

\paradot{Examples}
In full-information {\em games} (like chess) with static opponent,
it is sufficient to know the current state of the game
(board configuration) to play well (the history plays no role),
hence $\Phi(h_t)=o_t$ is a sufficient summary (Markov condition).
Classical {\em physics} is essentially predictable from position and
velocity of objects at a single time, or equivalently from the
locations at two consecutive times, hence $\Phi(h_t)=o_t o_{t-1}$ is
a sufficient summary (2nd order Markov).
For {\em i.i.d.\ processes} of unknown probability (e.g.\ clinical trials
$\simeq$ Bandits), the frequency of observations
$\Phi(h_n)=(\sum_{t=1}^n\delta_{o_t o})_{o\in\O}$
is a sufficient statistic.
In a {\em POMDP planning} problem, the so-called belief vector at
time $t$ can be written down explicitly as some function of the complete
history $h_t$ (by integrating out the hidden states). $\Phi(h_t)$
could be chosen as (a discretized version of) this belief vector, showing
that $\Phi$MDP generalizes POMDPs.
Obviously, the {\em identity} $\Phi(h)=h$ is always sufficient but not
very useful, since \Env() as a function of $\H$ is hard to
impossible to ``learn''. 

This suggests to look for $\Phi$ with small codomain, which allow
to learn/estimate/approximate $\Env$ by $\widehat\Env$ such that
$o_t\approx\widehat\Env(\Phi(h_{t-1}))$ for $t=1...n$.

\paradot{Example}
Consider a robot equipped with a camera, i.e.\ $o$ is a pixel image.
Computer vision algorithms usually extract a set of features from
$o_{t-1}$ (or $h_{t-1}$), from low-level patterns to high-level
objects with their spatial relation. Neither is it possible nor
necessary to make a precise prediction of $o_t$ from summary
$\Phi(h_{t-1})$. An approximate prediction must and will do.
The difficulty is that the similarity measure ``$\approx$'' needs to
be context dependent. Minor image nuances are irrelevant when
driving a car, but when buying a painting it makes a huge difference
in price whether it's an original or a copy. Essentially only a
bijection $\Phi$ would be able to extract {\em all potentially}
interesting features, but such a $\Phi$ defeats its original
purpose.

\paradot{From histories to states}
It is of utmost importance to properly formalize the meaning of
``$\approx$'' in a general, domain-independent way.
Let $s_t:=\Phi(h_t)$ summarize all relevant information in history
$h_t$. I call $s$ a state or feature (vector) of $h$. ``Relevant''
means that the future is predictable from $s_t$ (and $a_t$)
alone, and that the relevant future is coded in $s_{t+1}s_{t+2}...$.
So we pass from the complete (and known) history $o_1 a_1 r_1...o_n
a_n r_n$ to a ``compressed'' history $sar_{1:n}\equiv s_1 a_1 r_1...s_n
a_n r_n$ and seek $\Phi$ such that $s_{t+1}$ is (approximately a
stochastic) function of $s_t$ (and $a_t$).
Since the goal of the agent is to maximize his rewards, the rewards
$r_t$ are always relevant, so they (have to) stay untouched (this
will become clearer below).

\paradot{The $\Phi$MDP}
The structure derived above is a classical Markov Decision Process
(MDP), but the primary question I ask is not the usual one of
finding the value function or best action or comparing different
models of a given state sequence. I ask how well can the
state-action-reward sequence generated by $\Phi$ be modeled as an
MDP compared to other sequences resulting from different $\Phi$.

\section{$\mathbf\Phi$MDP Coding and Evaluation}\label{secCE}

I first review optimal codes and model selection methods for i.i.d.\
sequences, subsequently adapt them to our situation, and show that
they are suitable in our context. I state my Cost function for
$\Phi$ and the $\Phi$ selection principle.

\paradot{I.i.d.\ processes}
Consider i.i.d.\ $x_1...x_n\in\X^n$ for finite $\X=\{1,...,m\}$.
For known $\t_i=\P[x_t=i]$ we have
$\P(x_{1:n}|\v\t)=\t_{x_1}\cdot...\cdot\t_{x_n}$.
It is well-known that there exists a code (e.g.\ arithmetic or
Shannon-Fano) for $x_{1:n}$ of length $-\log\P(x_{1:n}|\v\t)$, which is
asymptotically optimal with probability one.

For unknown $\v\t$ we may use a frequency estimate $\hat\t_i=n_i/n$,
where $n_i=|\{t:x_t=i\}|$. Then
$-\log\P(x_{1:n}|\v{\hat\t})=n\,H(\v{\hat\t})$, where
$H(\v{\hat\t}):=-\sum_{i=1}^{m}\hat\t_i\log\hat\t_i$ is the Entropy
of $\v{\hat\t}$ ($0\log 0:=0=:0\log{0\over 0}$). We also need to
code $(n_i)$, which naively needs $\log n$ bits for each $i$. One
can show that it is sufficient to code each $\hat\t_i$ to accuracy
$O(1/\sqrt{n})$, which requires only $\fr12\log n+O(1)$ bits each.
Hence the overall code length of $x_{1:n}$ for unknown frequencies
is
\beq\label{iidCodeL}
  \CL(x_{1:n}) \;=\; \CLN(\v n) \;:=\;
  n\,H(\v n/n) + \fr{m'-1}2\log n
  \iftwocol\else\qmbox{for} n>0 \qmbox{and} 0 \qmbox{else.}\fi
\eeq
\iftwocol for $n>0$ and 0 else, \fi
where $\v n=(n_1,...,n_m)$ and $n=n_+=n_1+...+n_m$ and
$m'=|\{i:n_i>0\}|\leq m$ is the number of non-empty categories.
The code is optimal (within $+O(1)$) for all i.i.d.\ sources.
It can be rigorously derived from many principles: %
MDL, MML, combinatorial, incremental, and Bayesian \cite{Gruenwald:07book}.

In the following I will ignore the $O(1)$ terms and refer to
\req{iidCodeL} simply as {\em the} code length. Note that $x_{1:n}$
is coded exactly (lossless). Similarly (see MDP below) sampling
models more complex than i.i.d.\ may be considered, and the one that
leads to the shortest code is selected as the best model
\cite{Gruenwald:07book}.

\paradot{MDP definitions}
Recall that a sequence $sar_{1:n}$ is said to be sampled from an MDP
$(\S,\A,T,R)$ iff the probability of $s_t$ only depends on $s_{t-1}$
and $a_{t-1}$; and $r_t$ only on
$s_{t-1}$, $a_{t-1}$, and $s_t$. That is,
\bqan
  \P(s_t|h_{t-1}a_{t-1}) \;=\; \P(s_t|s_{t-1},a_{t-1}) &=:& T_{s_{t-1}s_t}^{a_{t-1}} \\
         \P(r_t|h_t) \;=\; \P(r_t|s_{t-1},a_{t-1},s_t) &=:& R_{s_{t-1}s_t}^{a_{t-1}r_t}
\eqan
For simplicity of exposition I assume a deterministic dependence of
$r_t$ on $s_t$ only, i.e.\ $r_t=R_{s_t}$.
In our case, we can identify the state-space $\S$ with the states
$s_1,...,s_n$ ``observed'' so far. Hence $\S=\{s^1,...,s^m\}$ is
finite and typically $m\ll n$, i.e.\ states repeat. Let
$s\stackrel{a}\to s'(r')$ be shorthand for ``action $a$ in state $s$
resulted in state $s'$ (reward $r'$)''. Let $\T_{ss'}^{ar'}:=\{t\leq
n:s_{t-1}=s, a_{t-1}=a, s_t=s', r_t=r'\}$ be the set of times $t-1$
at which $s\stackrel{a}\to s'r'$, and
$n_{ss'}^{ar'}:=|\T_{ss'}^{ar'}|$ their number
($n_{\p\p}^{\p\p}=n$).

\paradot{Coding MDP sequences}
For some fixed $s$ and $a$, consider the subsequence
$s_{t_1}...s_{t_{n'}}$ of states reached from $s$ via $a$
($s\stackrel{a}\to s_{t_i}$), i.e.\
$\{t_1,...,t_{n'}\}=\T_{s*}^{a*}$, where $n'=n_{s\p}^{a\p}$. By
definition of an MDP, this sequence is i.i.d.\ with $s'$ occurring
$n'_{s'}:=n_{ss'}^{a\p}$ times. By \req{iidCodeL} we can code this
sequence in $\CL(\v n')$ bits. The whole sequence $s_{1:n}$ consists
of $|\S\times\A|$ i.i.d.\ sequences, one for each
$(s,a)\in\S\times\A$. We can join their codes and get a total code
length
\beq\label{sCode}
  \CL(s_{1:n}|a_{1:n}) \;=\; \sum_{s,a} \CLN(\v n_{s\vi}^{a\p})
\eeq
Similarly to the states we code the rewards. There are different
``standard'' reward models. I consider only the simplest case of a
small discrete reward set $\R$ like $\{0,1\}$ or $\{-1,0,+1\}$ here
and defer generalizations to $\SetR$ and a discussion of variants to
the $\Phi$DBN model \cite{Hutter:09phidbn}.
By the MDP assumption, for each state $s'$, the rewards at times
$\T_{\p s'}^{\p *}$ are i.i.d. Hence they can be coded in
\beq\label{rCodeEx}
  \CL(r_{1:n}|s_{1:n},a_{1:n}) \;=\; \sum_{s'} \CLN(\v n_{\p s'}^{\p\vi})
\eeq
bits. I have been careful to assign zero code length to
non-occurring transitions $s\stackrel{a}\to s'r'$ so that large but
sparse MDPs don't get penalized too much.

\paradot{Reward$\mathbf\leftrightarrow$state trade-off}
Note that the code for $\v r$ depends on $\v s$. Indeed we may
interpret the construction as follows: Ultimately we/the agent cares
about the reward, so we want to measure how well we can predict the
rewards, which we do with\req{rCodeEx}.
But this code depends on $\v s$, so we need a code for $\v s$ too,
which is \req{sCode}. To see that we need both parts consider two
extremes.

A simplistic state transition model (small $|\S|$) results in a
short code for $\v s$. For instance, for $|\S|=1$, nothing needs to be
coded and \req{sCode} is identically zero. But this obscures
potential structure in the reward sequence, leading to a long code
for $\v r$.

On the other hand, the more detailed the state transition model
(large $|\S|$) the easier it is to predict and hence compress $\v
r$. But a large model is hard to learn, i.e.\ the code for $\v s$
will be large. For instance for $\Phi(h)=h$, no state repeats and
the frequency-based coding breaks down.

\paradot{$\Phi$ selection principle}
Let us define the {\em Cost} of $\Phi:\H\to\S$ on $h_n$ as the
length of the $\Phi$MDP code for $\v s\v r$ given $\v a$:
\bqa\label{costphi}
  & & \nq\nq\Cost(\Phi|h_n) \;:=\;
      \CL(s_{1:n}|a_{1:n}) + \CL(r_{1:n}|s_{1:n},a_{1:n}),
\\ \nonumber
  & & \nq\nq\qmbox{where} s_t=\Phi(h_t)\qmbox{and} h_t=oar_{1:t-1}o_t
\eqa
The discussion above suggests that the minimum of the joint code
length, i.e.\ the Cost, is attained for a $\Phi$ that keeps all and
only relevant information for predicting rewards. Such a $\Phi$ may
be regarded as best explaining the rewards. So we are looking for a
$\Phi$ of minimal cost:
\beq\label{bestphi}
   \Phi^{best} \;:=\; \arg\min_\Phi\{ \Cost(\Phi|h_n) \}
\eeq
The state sequence generated by $\Phi^{best}$ (or approximations
thereof) will usually only be approximately MDP. While
$\Cost(\Phi|h)$ is an optimal code only for MDP sequences, it still
yields good codes for approximate MDP sequences. Indeed,
$\Phi^{best}$ balances closeness to MDP with simplicity. The primary
purpose of the simplicity bias is {\em not} computational
tractability, but generalization ability
\cite{Hutter:07iorx,Hutter:04uaibook}.

\section{A Tiny Example}\label{secTE}

The purpose of the tiny example in this section is to provide enough
insight into how and why $\Phi$MDP works to convince the reader that
our $\Phi$ selection principle is reasonable.
Consider binary observation space $\O=\{0,1\}$, quaternary reward
space $\R=\{0,1,2,3\}$, and a single action $\A=\{0\}$. Observations
$o_t$ are independent fair coin flips, i.e.\ Bernoulli($\fr12$), and
reward $r_t=2o_{t-1}+o_t$ a deterministic function of the two most
recent observations.

\paradot{Considered features}
As features $\Phi$ I consider $\Phi_k:\H\to\O^k$ with
$\Phi_k(h_t)=o_{t-k+1}...o_t$ for various $k=0,1,2,...$ which regard
the last $k$ observations as ``relevant''.
Intuitively $\Phi_2$ is the best observation summary, which I confirm
below. The state space $\S=\{0,1\}^k$ (for sufficiently large $n$).
The $\Phi$MDPs for $k=0,1,2$ are as follows.

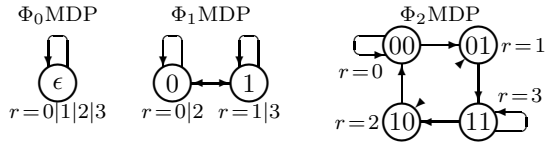
\begin{center}
\iftwocol\footnotesize\unitlength=2.1ex\else\small\unitlength=3ex\fi
\linethickness{0.4pt}
\begin{picture}(27,7)(0,0.3)
\thinlines
\put(2,7){\makebox(0,0)[ct]{$\Phi_0$MDP}}
\thicklines\put(2,3){\circle{2}}\thinlines
\put(2,3){\makebox(0,0)[cc]{\normalsize$\epstr$}}
\put(2,2){\makebox(0,0)[ct]{$r=0|1|2|3$}}
\put(2,5){\oval(1,1)[t]}
\put(1.5,5){\vector(0,-1){1.13}}
\put(2.5,5){\line(0,-1){1.13}}

\put(10,7){\makebox(0,0)[ct]{$\Phi_1$MDP}}
\thicklines\put(8,3){\circle{2}}\thinlines
\put(8,3){\makebox(0,0)[cc]{\normalsize 0}}
\put(8,2){\makebox(0,0)[ct]{$r=0|2$}}
\put(8,5){\oval(1,1)[t]}
\put(7.5,5){\vector(0,-1){1.13}}
\put(8.5,5){\line(0,-1){1.13}}
\put(10,3){\vector(1,0){1}}
\put(10,3){\vector(-1,0){1}}
\thicklines\put(12,3){\circle{2}}\thinlines
\put(12,3){\makebox(0,0)[cc]{\normalsize 1}}
\put(12,2){\makebox(0,0)[ct]{$r=1|3$}}
\put(12,5){\oval(1,1)[t]}
\put(11.5,5){\vector(0,-1){1.13}}
\put(12.5,5){\line(0,-1){1.13}}

\put(22,7){\makebox(0,0)[ct]{$\Phi_2$MDP}}
\thicklines\put(20,5){\circle{2}}\thinlines
\put(20,5){\makebox(0,0)[cc]{\normalsize 00}}
\put(19,4){\makebox(0,0)[rt]{$r=0$}}
\put(18,5){\oval(1,1)[l]}
\put(18,4.5){\vector(1,0){1.13}}
\put(18,5.5){\line(1,0){1.13}}
\thicklines\put(24,1){\circle{2}}\thinlines
\put(24,1){\makebox(0,0)[cc]{\normalsize 11}}
\put(25,2){\makebox(0,0)[lb]{$r=3$}}
\put(26,1){\oval(1,1)[r]}
\put(26,1.5){\vector(-1,0){1.13}}
\put(26,0.5){\line(-1,0){1.13}}
\thicklines\put(24,5){\circle{2}}\thinlines
\put(24,5){\makebox(0,0)[cc]{\normalsize 01}}
\put(25,5){\makebox(0,0)[lc]{$\,r=1$}}
\thicklines\put(20,1){\circle{2}}\thinlines
\put(20,1){\makebox(0,0)[cc]{\normalsize 10}}
\put(19,1){\makebox(0,0)[rc]{$\,r=2\,$}}
\put(21,5){\vector(1,0){2}}
\put(24,4){\vector(0,-1){2}}
\put(23,1){\vector(-1,0){2}}
\put(20,2){\vector(0,1){2}}
\put(22,3){\vector(1,1){1.3}}
\put(22,3){\vector(-1,-1){1.3}}
\end{picture}
\end{center}

\paranodot{$\mathbf\Phi_2$MDP}
with all non-zero transition probabilities being 50\% is an exact
representation of our data source. The missing arrow (directions)
are due to the fact that $s=o_{t-1}o_t$ can only lead to $s'=o'_t
o'_{t+1}$ for which $o'_t=o_t$. Note that $\Phi$MDP does not
``know'' this and has to learn the (non)zero transition
probabilities. Each state has two successor states with equal
probability, hence generates (see previous paragraph) a
Bernoulli($\fr12$) state subsequence and a constant reward sequence,
since the reward can be computed from the state = last two
observations. Asymptotically, all four states occur equally often,
hence the sequences have approximately the same length $n/4$.

In general, if $\v s$ (and similarly $\v r$) consists of
$x\in\SetN$ i.i.d.\ subsequences of equal length $n/x$ over $y\in\SetN$
symbols, the code length \req{sCode} (and similarly \req{rCodeEx}) is
\bqan
  \CL(\v s|\v a;x_y)   &=& \textstyle n\log y + x{|\S|-1\over 2}\log{n\over x}, \\
  \CL(\v r|\v s,\v a;x_y) &=& \textstyle n\log y + x{|\R|-1\over 2}\log{n\over x}
\eqan
where the extra argument $x_y$ just indicates the sequence property.
So for $\Phi_2$MDP we get
\beqn
  \CL(\v s|\v a;4_2) = n+6\log\fr n4 \qmbox{and}
  \CL(\v r|\v s,\v a;4_1) = 6\log\fr n4
\eeqn
The log-terms reflect the required memory to code (or the time to
learn) the MDP structure and probabilities. Since each state has
only 2 realized/possible successors, we need $n$ bits to code the
state sequence. The reward is a deterministic function of the state,
hence needs no memory to code given $\v s$.

\paranodot{The $\mathbf\Phi_0$MDP}
throws away all observations (left figure above), hence
$\CL(\v s|\v a;1_1)=0$. While the reward sequence is {\em not} i.i.d.\
(e.g.\ $r_{t+1}=3$ cannot follow $r_t=0$), $\Phi_0$MDP has no choice
regarding them as i.i.d., resulting in $\CL(\v s|\v a;1_4)=2n+\fr32\log
n$.

\paranodot{The $\mathbf\Phi_1$MDP}
model is an interesting compromise (middle figure above). The state
allows a partial prediction of the reward: State 0 allows rewards 0
and 2; state 1 allows rewards 1 and 3. Each of the two states
creates a Bernoulli($\fr12$) state successor subsequence and a
binary reward sequence, wrongly presumed to be Bernoulli($\fr12$).
Hence $\CL(\v s|\v a;2_2)=n+\log\fr n2$ and $\CL(\v r|\v s,\v
a;2_2)=n+3\log\fr n2$.

\paradot{Summary}
The following table summarizes the results for general $k=0,1,2$ and beyond:
\beqn\arraycolsep3pt\textmuskip
\begin{array}{c|c|c|c}
  \Cost(\Phi_0|h) & \Cost(\Phi_1|h) & \Cost(\Phi_2|h) & \Cost(\Phi_{k\geq 2}|h) \\ \hline
  2n+\fr32\log n & 2n+4\log\fr n2 & n+12\log\fr n4 & n+{2^k+2\over 2^{1-k}}\log\!\fr{n}{2^k}
\end{array}
\eeqn
For large $n$, $\Phi_2$ results in the shortest
code, as anticipated. The ``approximate'' model $\Phi_1$ is just not
good enough to beat the vacuous model $\Phi_0$, but in more
realistic examples some approximate model usually has the shortest
code.
In \cite{Hutter:09phidbn} I show on a more complex example how
$\Phi^{best}$ will store long-term information in a POMDP
environment.

\section{\Cost($\mathbf\Phi$) Minimization}\label{secCM}

I have reduced the reinforcement learning problem to a formal
$\Phi$-optimization problem. I briefly explain what we have gained
by this reduction, and provide some general information about
problem representations, stochastic search, and $\Phi$
neighborhoods. Finally I present a simplistic but concrete
algorithm for searching context tree MDPs.

\paradot{$\mathbf\Phi$ search}
I now discuss how to find good summaries $\Phi$. The introduced
generic cost function $\Cost(\Phi|h_n)$, based on only the known
history $h_n$, makes this a well-defined task that is completely
decoupled from the complex (ill-defined) reinforcement learning
objective. This reduction should not be under-estimated. We can
employ a wide range of optimizers and do not even have to worry
about overfitting. The most challenging task is to come up with
creative algorithms proposing $\Phi$'s.

There are many optimization methods: Most of them are search-based:
random, blind, informed, adaptive, local, global, population based,
exhaustive, heuristic, and other search methods \cite{Aarts:97}.
Most are or can be adapted to the structure of the objective
function, here $\Cost(\cdot|h_n)$. Some exploit the structure more
directly (e.g.\ gradient methods for convex functions). Only in very
simple cases can the minimum be found analytically (without search).

General maps $\Phi$ can be represented by/as programs for which
variants of Levin search \cite{Schmidhuber:04oops,Hutter:04uaibook}
and genetic programming are the major search algorithms.
Decision trees/lists/grids are also quite powerful, especially
rule-based ones in which logical expressions recursively divide
domain $\H$ into ``true/false'' regions \cite{Sanner:08} that can be
identified with different states.

\paradot{$\Phi$ neighborhood relation}
Most search algorithms require the specification of a neighborhood
relation or distance between candidate $\Phi$.
A natural ``minimal'' change of $\Phi$ is splitting and merging
states (state refinement and coarsening).
Let $\Phi'$ split some state $s^a\in\S$ of $\Phi$ into
$s^b,s^c\not\in\S$
\beqn
  \Phi'(h) \;:=\; \left\{ {\Phi(h) \quad\qmbox{if} \Phi(h)\neq s^a \atop
                           s^b \mbox{ or } s^c \qmbox{if} \Phi(h)=s^a} \right.
\eeqn
where the histories in state $s^a$ are distributed among $s^b$ and
$s^c$ according to some splitting rule (e.g.\ randomly). The new
state space is $\S'=\S\setminus\{s^a\}\cup\{s^b,s^c\}$.
Similarly $\Phi'$ merges states $s^b,s^c\in\S$ into $s^a\not\in\S$ if
\beqn
  \Phi'(h) \;:=\; \left\{ {\phi(h) \qmbox{if} \Phi(h)\neq s^a\qquad \atop
                           s^a \quad\;\qmbox{if} \Phi(h)=s^b \mbox{ or } s^c} \right.
\eeqn
where $\S'=\S\setminus\{s^b,s^c\}\cup\{s^s\}$. We can regard
$\Phi'$ as being a neighbor of or similar to $\Phi$.

\paradot{Stochastic $\Phi$ search}
Stochastic search is the method of choice for high-dimensional
unstructured problems. Monte Carlo methods can actually be highly
effective, despite their simplicity \cite{Liu:02}. The general idea
is to randomly choose a neighbor $\Phi'$ of $\Phi$ and replace
$\Phi$ by $\Phi'$ if it is better, i.e.\ has smaller Cost. Even if
$\Cost(\Phi'|h)>\Cost(\Phi|h)$ we may keep $\Phi'$, but only with
some (in the cost difference exponentially) small probability.
Simulated annealing is a version which minimizes $\Cost(\Phi|h)$.
Apparently, $\Phi$ of small cost are (much) more likely to occur
than high cost $\Phi$.

\paradot{Context tree example}
The $\Phi_k$ in Section \ref{secTE} depended on the last $k$
observations. Let us generalize this to a context dependent variable
length:
Consider a finite complete suffix free set of strings (= prefix tree
of reversed strings) $\S\subset\O^*$ as our state space (e.g.\
$\S=\{0,01,011,111\}$ for binary $\O$), and define $\Phi_\S(h_n):=s$
iff $o_{n-|s|+1:n}=s\in\S$, i.e.\ $s$ is the part of
the history regarded as relevant.
State splitting and merging works as follows: For binary $\O$, if
history part $s\in\S$ of $h_n$ is deemed too short, we replace $s$
by $0s$ and $1s$ in $\S$, i.e.\ $\S'=\S\setminus\{s\}\cup\{0s,1s\}$.
If histories $1s,0s\in\S$ are deemed too long, we replace them by
$s$, i.e.\ $\S'=\S\setminus\{0s,1s\}\cup\{s\}$.
Large $\O$ might be coded binary and then treated similarly. The
idea of using suffix trees as state space is from
\cite{McCallum:96}. For small $\O$ we have the following simple
$\Phi$-optimizer:

\def\algitsep{\itemsep=0ex}
\begin{list}{}{\parskip=0ex\parsep=0ex\algitsep\leftmargin=0ex\labelwidth=0ex}
  \item {\bf\boldmath $\Phi$Improve($\Phi_\S,h_n$)}
  \begin{list}{}{\parskip=0ex\parsep=0ex\algitsep\leftmargin=2ex\labelwidth=1ex\labelsep=1ex}
    \item[$\lceil$] Randomly choose a state $s\in\S$;
    \item Let $p$ and $q$ be uniform random numbers in $[0,1]$;
    \item if $(p>1/2)$ then split $s$ i.e.\ $S'=S\setminus\{s\}\cup\{os:o\in\O\}$
    \item else if $\{os:o\in\O\}\subseteq\S$
    \item then merge them, i.e.\ $S'=S\setminus\{os:o\in\O\}\cup\{s\}$;
    \item if $(\Cost(\Phi_\S|h_n)-\Cost(\Phi_{\S'}|h_n) > \log(q))$ then $\S:=\S'$;
    \item [$\lfloor$] {\bf\boldmath return ($\Phi_\S$); }
  \end{list}
\end{list}

\section{Exploration \& Exploitation}\label{secEE}

Having obtained a good estimate $\hat\Phi$ of $\Phi^{best}$ in the
previous section, we can/must now determine a good action for our
agent. For a finite MDP with known transition probabilities, finding
the optimal action is routine. For estimated probabilities we run
into the infamous exploration-exploitation problem, for which
promising approximate solutions have recently been suggested
\cite{Szita:08}. At the end of this section I present the overall
algorithm for our $\Phi$MDP agent.

\paradot{Optimal actions for known MDPs}
For a known finite MDP $(\S,\A,T,R,\gamma)$,
the maximal achievable (``optimal'') expected future discounted reward sum,
called ($Q$) $V\!$alue (of action $a$) in state
$s$, satisfies the following (Bellman) equations \cite{Sutton:98}
\beq\label{BellmanEq}
  Q_s^{*a} \;=\; \sum_{s'} T_{ss'}^a[R_{ss'}^a + \gamma V_{s'}^*]
  \qmbox{and} V_s^*=\max_a Q_s^{*a}
\eeq
where $0<\g<1$ is a discount parameter, typically close to 1. See
\cite[Sec.5.7]{Hutter:04uaibook} for proper choices. The equations
can be solved in polynomial time by a simple iteration process or
various other methods \cite{Puterman:94}.
After observing $o_{n+1}$, the optimal next action is
\beq\label{BellmanSol}
  a_{n+1} := \arg\max_a Q_{s_{n+1}}^{*a},\qmbox{where} s_{n+1}=\Phi(h_{n+1})
\eeq

\paradot{Estimating the MDP}
We can estimate the transition probability $T$ by
\beq\label{hatT}
  \hat T_{ss'}^a \;:=\; {n_{ss'}^{a\p}\over n_{s\p}^{a\p}}
  \qmbox{if} n_{s\p}^{a\p}>0 \qmbox{and} 0 \qmbox{else.}
\eeq
It is easy to see that the Shannon-Fano code of $s_{1:n}$ based on
$\P_{\smash{\!\hat T}}(s_{1:n}|a_{1:n})=\prod_{t=1}^n\hat
T_{s_{t-1}s_t}^{a_{t-1}}$ plus the code of the non-zero transition
probabilities $\hat T_{ss'}^a>0$ to relevant accuracy
$O(1/\sqrt{n_{s\p}^{a\p}})$ has length \req{sCode}, i.e.\
the frequency estimate \req{hatT} is consistent with the
attributed code length.
The expected reward can be estimated as

\beq\label{hatR}
  \hat R_{ss'}^a := \sum_{r'\in\R}\hat R_{ss'}^{ar'} r',\qquad
  \hat R_{ss'}^{ar'} := {n_{ss'}^{ar'}\over n_{ss'}^{a\p}}
\eeq

\paradot{Exploration}
Simply replacing $T$ and $R$ in \req{BellmanEq} and \req{BellmanSol}
by their estimates \req{hatT} and \req{hatR} can lead
to very poor behavior, since parts of the state space may never be
explored, causing the estimates to stay poor.

Estimate $\hat T$ improves with increasing $n_{s\p}^{a\p}$, which
can (only) be ensured by trying all actions $a$ in all states $s$
sufficiently often. But the greedy policy above has no incentive to
explore, which may cause the agent to perform very poorly: The agent
stays with what he {\em believes} to be optimal without trying to
solidify his belief. Trading off exploration versus exploitation
optimally is computationally intractable
\cite{Hutter:04uaibook,Poupart:06,Ross:08bayes} in all but extremely
simple cases (e.g.\ Bandits).
Recently, polynomially optimal algorithms (Rmax,E3,OIM) have been
invented \cite{Kearns:98,Szita:08}: An agent is more explorative if
he expects a high reward in the unexplored regions. We can
``deceive'' the agent to believe this by adding another
``absorbing'' high-reward state $s^e$ to $\S$, not in the range of
$\Phi(h)$, i.e.\ never observed. Henceforth, $\S$ denotes the
extended state space. For instance $+$ in \req{hatT} now includes
$s^e$. We set
\beq\label{extnR}
  n_{ss^e}^a=1,\quad n_{s^e s}^a=\delta_{s^e s},\quad R_{ss^e}^a=R_{max}^e
\eeq
for all $s,a$, where exploration bonus $R_{max}^e$ is polynomially (in
$(1-\g)^{-1}$ and $|\S\times\A|$) larger than $\max\R$
\cite{Szita:08}.

Now compute the agent's action by \req{BellmanEq}-\req{hatR} but for
the extended $\S$. The optimal policy $p^*$ tries to find a chain of
actions and states that likely leads to the high reward absorbing
state $s^e$. Transition $\hat T_{ss^e}^a=1/n_{s+}^a$ is only ``large'' for
small $n_{s+}^a$, hence $p^*$ has a bias towards unexplored
(state,action) regions. It can be shown that this algorithm makes
only a polynomial number of sub-optimal actions.

The overall algorithm for our $\Phi$MDP agent is as follows.

\def\algitsep{\itemsep=0ex}
\begin{list}{}{\parskip=0ex\parsep=0ex\algitsep\leftmargin=0ex\labelwidth=0ex}
  \item {\bf\boldmath $\Phi$MDP-Agent($\A,\R$)}
  \begin{list}{}{\parskip=0ex\parsep=0ex\algitsep\leftmargin=2ex\labelwidth=1ex\labelsep=1ex}
    \item[$\lceil$] Initialize $\Phi\equiv\epstr$; $\;\S=\{\epstr\}$; $\;h_0=a_0=r_0=\epstr$;
    \item for $n=0,1,2,3,...$
    \begin{list}{}{\parskip=0ex\parsep=0ex\algitsep\leftmargin=2ex\labelwidth=1ex\labelsep=1ex}
      \item[$\lceil$] Choose e.g.\ $\g=1-1/(n+1)$;
      \item Set $R_{max}^e=$Polynomial$((1-\g)^{-1},|\S\times\A|)\cdot\max\R$;
      \item While waiting for $o_{n+1}$ \iftwocol\else (and $r_{n+1}$ below) \fi $\{\Phi:=\Phi$Improve($\Phi,h_n$)$\}$;
      \item Observe $o_{n+1}$; $\;h_{n+1}=h_n a_n r_n o_{n+1}$;
      \item $s_{n+1}:=\Phi(h_{n+1})$; $\;\S:=\S\cup\{s_{n+1}\}$;
      \item Compute action $a_{n+1}$ from Equations \req{BellmanEq}-\req{extnR};
      \item Output action $a_{n+1}$;
    \end{list}
    \item [$\lfloor$] $\lfloor$ Observe reward $r_{n+1}$;
  \end{list}
\end{list}

\section{Improved Cost Function}\label{secICF}

As discussed, we ultimately only care about (modeling) the rewards,
but this endeavor required introducing and coding states. The
resulted Cost($\Phi|h$) function is a code length of not only the
rewards but also the ``spurious'' states. This likely leads to a too
strong penalty of models $\Phi$ with large state spaces $\S$. The
proper Bayesian formulation developed in this section allows to
``integrate'' out the states. This leads to a code for the rewards
only, which better trades off accuracy of the reward model and state
space size.

For an MDP with transition and reward probabilities $T_{ss'}^a$ and
$R_{ss'}^{ar'}$, the probabilities of the state and reward sequences
are
\beqn
  \P\iftwocol\else _T\fi(s_{1:n}|a_{1:n}) = \!\prod_{t=1}^n\! T_{s_{t-1}s_t}^{a_{t-1}},\quad\!\!
  \P\iftwocol\else _R\fi(r_{1:n}|s_{1:n}a_{1:n}) = \!\prod_{t=1}^n\! R_{s_{t-1}s_t}^{a_{t-1}r_t}
\eeqn
The probability of $\v r|\v a$ can be obtained by taking the product
and marginalizing $\v s$:
\beqn
  \P_{\!U}(r_{1:n}|a_{1:n})
  \iftwocol\else = \sum_{s_{1:n}}\P_{\!T}(s_{1:n}|a_{1:n}) \P_{\!R}(r_{1:n}|s_{1:n}a_{1:n}) \fi
  =\!\! \sum_{s_{1:n}}\prod_{t=1}^n U_{s_{t-1}s_t}^{a_{t-1}r_t}
  =\!\! \sum_{s_n}[U^{a_0 r_1}\!\cdot\cdot\cdot U^{a_{n-1}r_n}]_{s_0s_n}
\eeqn
where for each $a\in\A$ and $r'\in\R$, matrix
$U^{ar'}\in\SetR^{m\times m}$ is defined as $[U^{ar'}]_{ss'}\equiv
U_{ss'}^{ar'}:=T_{ss'}^a R_{ss'}^{ar'}$. The right $n$-fold matrix
product can be evaluated in time $O(m^2 n)$. This shows that
$\v r$ given $\v a$ and $U$ can be coded in $-\log P_U$ bits.
The unknown $U$ needs to be estimated, e.g.\ by the relative
frequency $\hat U_{ss'}^{ar'}:=n_{ss'}^{ar'}/n_{s\p}^{a\p}$. The
$M:=m(m-1)|\A|(|\R|-1)$ (independent) elements of $\hat U$ can be
coded to sufficient accuracy in $\fr12 M\log n$ bits. Together this
leads to a code for $\v r|\v a$ of length
\beq\label{ICost}
  \mbox{ICost}(\Phi|h_n) \;:=\; - \log\P_{\!\hat U}(r_{1:n}|a_{1:n}) + \fr12 M\log n
\eeq
In practice, $M$ can and should be chosen smaller like done in the
original \Cost\ function, where we have used a restrictive model for
$R$ and considered only non-zero transitions in $T$.

\section{Conclusion}\label{secDisc}

I have developed a formal criterion for evaluating and selecting
good ``feature'' maps $\Phi$ from histories to states and
presented the feature reinforcement learning algorithm
$\Phi$MDP-Agent(). The computational flow is
$h\leadsto\hat\Phi\leadsto(\hat T,\hat R)\leadsto(\hat V,\hat
Q)\leadsto a$.
%
The algorithm can easily and significantly be accelerated:
Local search algorithms produce sequences of ``similar'' $\Phi$,
which naturally suggests to compute/update $\Cost(\Phi|h)$ and the
value function $V$ incrementally.
%
The primary purpose of this work was to introduce and explore
$\Phi$-selection on the conveniently simple (but impractical)
unstructured finite MDPs. The results of this work set the stage for
the more powerful $\Phi$DBN model developed in the companion article
\cite{Hutter:09phidbn} based on Dynamic Bayesian Networks.
%
The major open problems are to develop smart $\Phi$ generation and
smart stochastic search algorithms for $\Phi^{best}$, and to determine
whether minimizing \req{ICost} is the right criterion.


\begin{small}
\newcommand{\etalchar}[1]{$^{#1}$}

\end{small}


\begin{thebibliography}{ABCD}

\bibitem[AL97]{Aarts:97}
E.~H.~L. Aarts and J.~K. Lenstra, editors.
\newblock {\em Local Search in Combinatorial Optimization}.
\newblock Discrete Mathematics and Optimization. Wiley-Interscience,
  Chichester, England, 1997.

\bibitem[GDG03]{Givan:03}
R.~Givan, T.~Dean, and M.~Greig.
\newblock Equivalence notions and model minimization in {M}arkov decision
  processes.
\newblock {\em Artificial Intelligence}, 147(1--2):163--223, 2003.

\bibitem[Gor99]{Gordon:99}
G.~Gordon.
\newblock {\em Approximate Solutions to {M}arkov Decision Processes}.
\newblock PhD thesis, School of Computer Science, Carnegie Mellon University,
  Pittsburgh, PA, 1999.

\bibitem[GP07]{Goertzel:07}
B.~Goertzel and C.~Pennachin, editors.
\newblock {\em Artificial General Intelligence}.
\newblock Springer, 2007.

\bibitem[Gr{\"u}07]{Gruenwald:07book}
P.~D. Gr{\"u}nwald.
\newblock {\em The Minimum Description Length Principle}.
\newblock The MIT Press, Cambridge, 2007.

\bibitem[HTF01]{Hastie:01}
T.~Hastie, R.~Tibshirani, and J.~H. Friedman.
\newblock {\em The Elements of Statistical Learning}.
\newblock Springer, 2001.

\bibitem[Hut05]{Hutter:04uaibook}
M.~Hutter.
\newblock {\em Universal Artificial Intelligence: Sequential Decisions based on
  Algorithmic Probability}.
\newblock Springer, Berlin, 2005.
\newblock 300 pages, http://www.hutter1.net/ai/uaibook.htm.

\bibitem[Hut07]{Hutter:07aixigentle}
M.~Hutter.
\newblock Universal algorithmic intelligence: A mathematical
  top$\rightarrow$down approach.
\newblock In {\em Artificial General Intelligence}, pages 227--290. Springer,
  Berlin, 2007.

\bibitem[Hut09]{Hutter:09phidbn}
M.~Hutter.
\newblock Feature dynamic {B}ayesian networks.
\newblock In {\em Artificial General Intelligence ({AGI'09})}. Atlantis Press,
  2009.

\bibitem[KLC98]{Kaelbling:98}
L.~P. Kaelbling, M.~L. Littman, and A.~R. Cassandra.
\newblock Planning and acting in partially observable stochastic domains.
\newblock {\em Artificial Intelligence}, 101:99--134, 1998.

\bibitem[KS98]{Kearns:98}
M.~J. Kearns and S.~Singh.
\newblock Near-optimal reinforcement learning in polynomial time.
\newblock In {\em Proc. 15th International Conf. on Machine Learning}, pages
  260--268. Morgan Kaufmann, San Francisco, CA, 1998.

\bibitem[LH07]{Hutter:07iorx}
S.~Legg and M.~Hutter.
\newblock Universal intelligence: A definition of machine intelligence.
\newblock {\em Minds \& Machines}, 17(4):391--444, 2007.

\bibitem[Liu02]{Liu:02}
J.~S. Liu.
\newblock {\em Monte Carlo Strategies in Scientific Computing}.
\newblock Springer, 2002.

\bibitem[McC96]{McCallum:96}
A.~K. McCallum.
\newblock {\em Reinforcement Learning with Selective Perception and Hidden
  State}.
\newblock PhD thesis, Department of Computer Science, University of Rochester,
  1996.

\bibitem[Put94]{Puterman:94}
M.~L. Puterman.
\newblock {\em Markov Decision Processes --- Discrete Stochastic Dynamic
  Programming}.
\newblock Wiley, New York, NY, 1994.

\bibitem[PVHR06]{Poupart:06}
P.~Poupart, N.~A. Vlassis, J.~Hoey, and K.~Regan.
\newblock An analytic solution to discrete {B}ayesian reinforcement learning.
\newblock In {\em Proc. 23rd International Conf. on Machine Learning
  ({ICML'06})}, volume 148, pages 697--704, Pittsburgh, PA, 2006. ACM.

\bibitem[RN03]{Russell:03}
S.~J. Russell and P.~Norvig.
\newblock {\em Artificial Intelligence. {A} Modern Approach}.
\newblock Prentice-Hall, Englewood Cliffs, NJ, 2nd edition, 2003.

\bibitem[RP08]{Ross:08bayes}
S.~Ross and J.~Pineau.
\newblock Model-based {B}ayesian reinforcement learning in large structured
  domains.
\newblock In {\em Proc. 24th Conference in Uncertainty in Artificial
  Intelligence ({UAI'08})}, pages 476--483, Helsinki, 2008. AUAI Press.

\bibitem[RPPCd08]{Ross:08pomdp}
S.~Ross, J.~Pineau, S.~Paquet, and B.~Chaib-draa.
\newblock Online planning algorithms for {POMDP}s.
\newblock {\em Journal of Artificial Intelligence Research}, 2008(32):663--704,
  2008.

\bibitem[San08]{Sanner:08}
S.~Sanner.
\newblock {\em First-Order Decision-Theoretic Planning in Structured Relational
  Environments}.
\newblock PhD thesis, Department of Computer Science, University of Toronto,
  2008.

\bibitem[SB98]{Sutton:98}
R.~S. Sutton and A.~G. Barto.
\newblock {\em Reinforcement Learning: An Introduction}.
\newblock MIT Press, Cambridge, MA, 1998.

\bibitem[Sch04]{Schmidhuber:04oops}
J.~Schmidhuber.
\newblock Optimal ordered problem solver.
\newblock {\em Machine Learning}, 54(3):211--254, 2004.

\bibitem[SDL07]{Strehl:07}
A.~L. Strehl, C.~Diuk, and M.~L. Littman.
\newblock Efficient structure learning in factored-state {MDP}s.
\newblock In {\em Proc. 27th AAAI Conference on Artificial Intelligence}, pages
  645--650, Vancouver, BC, 2007. AAAI Press.

\bibitem[SL08]{Szita:08}
I.~Szita and A.~L{\"o}rincz.
\newblock The many faces of optimism: a unifying approach.
\newblock In {\em Proc. 12th International Conference ({ICML} 2008)}, volume
  307, Helsinki, Finland, June 2008.

\bibitem[SLJ{\etalchar{+}}03]{Singh:03}
S.~Singh, M.~Littman, N.~Jong, D.~Pardoe, and P.~Stone.
\newblock Learning predictive state representations.
\newblock In {\em Proc. 20th International Conference on Machine Learning
  ({ICML'03})}, pages 712--719, 2003.

\end{thebibliography}
\end{document}